\documentclass[a4paper,conference]{IEEEtran}
\ifCLASSINFOpdf
    \usepackage{graphicx}
    \usepackage{caption}
    \usepackage{amsmath}
\else
\fi
\graphicspath{{./images/}}
\newcommand{\hide}[1]{}

\newcommand{\etal}{\textit{et al.}~}

\begin{document}
%
\title{Learning Group Activities from Skeletons without Individual Action Labels }

\author{\IEEEauthorblockN{Fabio Zappardino, Tiberio Uricchio, Lorenzo Seidenari, Alberto del Bimbo}
\IEEEauthorblockA{University of Florence, Italy\\
\{name.surname\}@unifi.it} 
}


%


\maketitle

\begin{abstract}
To understand human behavior we must not just recognize individual actions but model possibly complex group activity and interactions. Hierarchical models obtain the best results in group activity recognition but require fine grained individual action annotations at the actor level. In this paper we show that using only skeletal data we can train a state-of-the art end-to-end system using only group activity labels at the sequence level.
Our experiments show that models trained without individual action supervision perform poorly. On the other hand we show that pseudo-labels can be computed from any pre-trained feature extractor with comparable final performance.
Finally our carefully designed lean pose only architecture shows highly competitive results versus more complex multimodal approaches even in the self-supervised variant.
\end{abstract}


%
\IEEEpeerreviewmaketitle

\section{Introduction}
Human behavior understanding can hardly be imagined as a task which requires to explain each individual actions in isolation. Human behavior is for the most part induced by social interactions.
For a machine to understand the meaning of multiple humans interacting with each other, so called social behavior or group activity, multiple level of reasoning must be enacted.
A naive approach could feed the whole frame to a deep convolutional neural network, but we know, that especially when domain shift is present that a large amount of data is required to learn a good hierarchy of features automatically. Recently hierarchical approaches have emerged. With such methodologies a model of the group behavior is built in a bottom-up fashion starting from the detection and tracking of all actors, following with the understanding of their individual behavior and finally  building collective models of the whole action.

As we know a fully supervised system is usually the best bet to obtain high accuracy. Unfortunately such systems must rely on a lot of hand labelling: each person action must be annotated at a not to low frequency, allowing a tracker to propagate such label over time. Considering this issue only few fully annotated datasets are available, limiting the development of group understanding computer vision algorithms. To address this issue in this work we propose a novel approach for group activity recognition based on the concept of pseudo-labels. Loosely inspired by the work of Caron \etal \cite{caron2018deep} we propose to replace costly single action labels with pseudo labels obtained via a simple clustering procedure which can be derived at very little cost. Interestingly, we show that such process is enough to provide such mid-level supervision thus enabling group activity recognition. Ground truth labels can therefore limited to the whole activity sequence with order of magnitude of time spared in the annotation phase.

Recently, the issue of privacy in A.I. applications has been raised especially in the EU, which enforces extremely strict policies regarding acquisition and protection of user personal information and data. Deploying cameras in public or private places to monitor user behavior has the major drawback of requiring the acquisition and possibly the storage and streaming of people images. While reliance on cryptography may offer a solution using highly anonymized human representation such as the skeleton offers many benefits. Human poses can be acquired in real-time with edge computing devices exploiting cloud computing facilities for the more complex task of action recognition. Moreover, dataset acquisition is made easier, not requiring the abidance to privacy policies if only the substantially anonymous 3D skeletal data is stored.

\begin{figure}[t]
    \includegraphics[width=\linewidth]{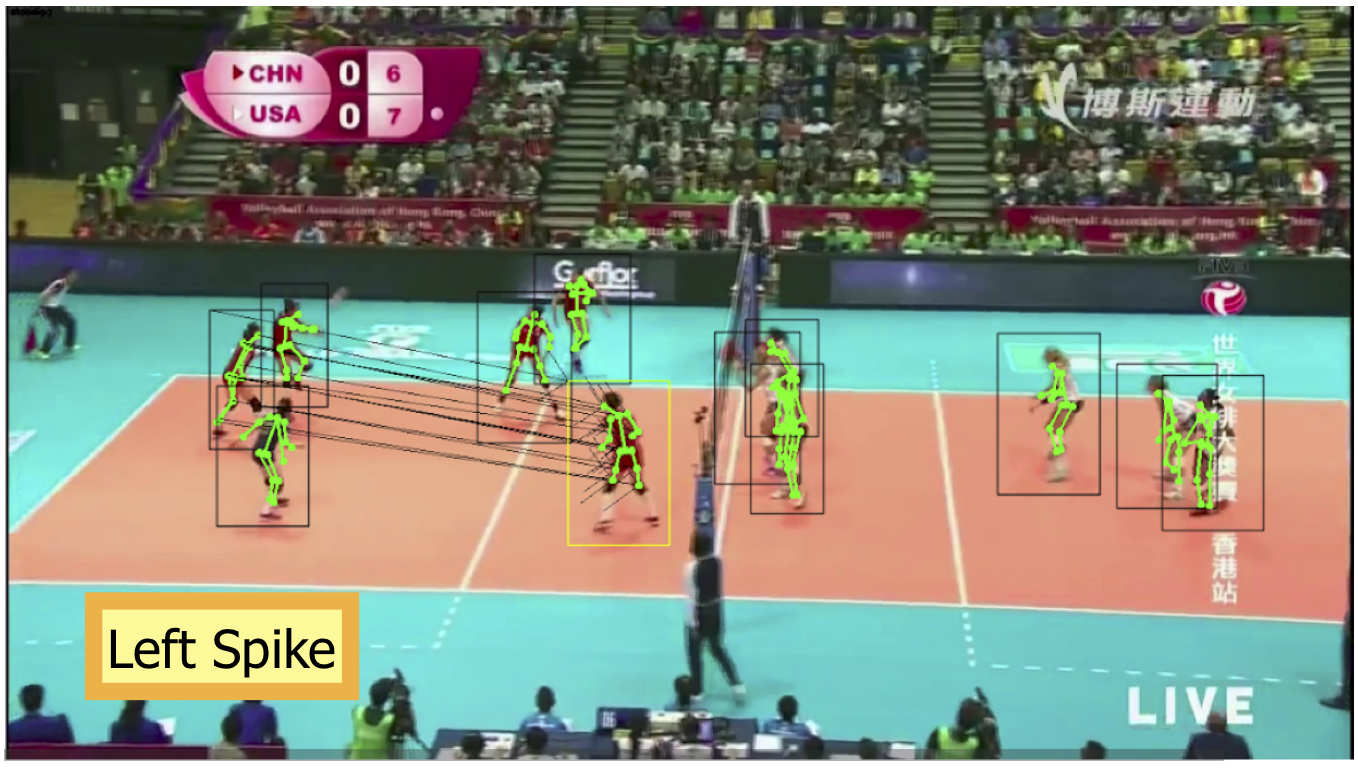}
	\caption{We classify group activities using skeletons, motion of single actors and their relative positions to a pivot actor.}
	\label{fig:eye}
\end{figure}
	
In this work we propose to relevant contribution to the field of activity recognition:
\begin{itemize}
	\item  We propose a novel semi-supervised approach allowing to train group activity recognition methods without fine grained ground truth annotation.
	\item We show how group activity can be efficiently performed using only skeletal representations which have a lower computational burden and have interesting privacy preserving properties.
\end{itemize}


\begin{figure*}
	
	\includegraphics[width=\linewidth]{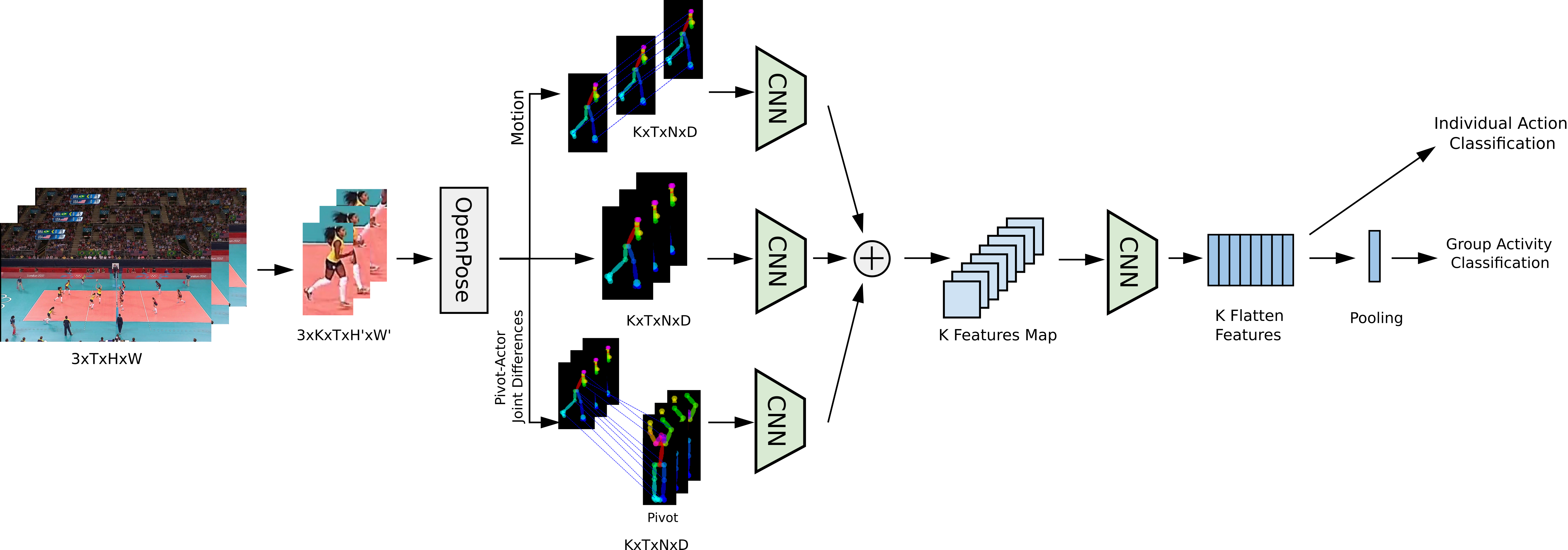}
	\caption{The proposed architecture employs three structurally identical branches with independent weights. We use openpose estimated skeletons to create three different representations. A concatenation of actor features is fed to a rather shallow network with two convolutional layers. Finally, single actor features are fed to the individual action classification head and single actor features are pooled to compute a group representation which is fed to the group classification head.}
	\label{fig:model}
\end{figure*}
\section{Related Works}
The initial methods for group activity recognition used handcrafted features which were extracted for each actor and combined using probabilistic graphical models \cite{amer2014hirf,choi2012unified,choi2013understanding,choi2011learning,hajimirsadeghi2015visual,lan2012social,lan2011discriminative}. Then, after the emergence of deep learning and the release of the Volleyball \cite{msibrahiCVPR16deepactivity} dataset, there was a renewed interest which remarkably increased performance. Most approaches use RNN-type networks which were introduced by Ibrahim \etal \cite{msibrahiCVPR16deepactivity} for the task of group activity recognition. They are trained to understand the action dynamics of individual actors and then predict group activity by taking their aggregation. A combination of RNN with graphical models were proposed in \cite{deng2016structure}. A two-level hierarchy of LSTMs were used in \cite{shu2017cern} to simultaneously minimize the loss of predictions and maximize the confidence. Bagautdinov \etal\cite{bagautdinov2017social} predicted every action of each actor and the group activity with an RNN which is also used to maintain temporal consistency of detected boxes. Intra-group, inter-group interactions and single person dynamics were considered in \cite{wang2017recurrent} and modeled with an LSTM. Each person is also modeled in a relational framework in \cite{ibrahim2018hierarchical}. Other works considered different approaches, Azar \etal \cite{azar2019convolutional} exploited activity maps of a CNN and iteratively refined group activity predictions. In \cite{wu2019learning}, they capture the appearance and positions between actors to build a graph of actor relationships from a CNN and graph convolutional networks. Gavrilyuk \etal \cite{gavrilyuk2020actor} explicitly modeled spatial and temporal relationships of actors with an actor-transformer model that learns and extract relevant information for group activity recognition. Our approach use CNN to perform classification. We share the use of skeletal data with \cite{gavrilyuk2020actor}, however our approach do not use additional 2D and 3D information to preserve privacy. We also consider the relative position of actors combined with motion and pose to perform individual action classification and group activity recognition. 

The use of human poses for recognizing actions of an actor is a popular approach in the literature. The early approaches were using handcrafted pose features \cite{xiaohan2015joint,wang2013approach}, then skeletons \cite{song2017end,liu2016spatio} and attention based pose estimations \cite{cao2016action,cheron2015p} were all explored for the task of action recognition of single actors. We exploit skeletons to model actions of single actors and then combine them into an holistic representation of the entire scene.

\section{Method}
In the following we show our approach for partially self-supervised group activity recognition using skeletal data.

\subsection{Input Skeleton Representation}
A group activity model has to consider individual actors  representation its temporal evolution and contextual information. Similarly to~\cite{msibrahiCVPR16deepactivity}, the person bounding boxes are firstly obtained through the object tracker in the Dlib library~\cite{article}. Then we feed each  person track frames as input to  openpose \cite{cao2018openpose}, obtaining a group of skeleton sequences $\mathbf{GS}$. $\mathbf{GS}$ can be represented with a $K \times T \times  N \times D$ tensor, where $K$ is the number of actors in the video clip, $T$ is the number of frames in the sequence, $N$ is the number of joints in the skeleton and $D$ is the coordinate dimension. Given a group of skeletons at time $t$, $\mathbf{GS}^t = \{\mathbf{S}^t_1, \mathbf{S}^t_2, \dots, \mathbf{S}^t_K\}$, we represent the skeleton of a person $k$ at time $t$ as 
\[\mathbf{S}^{tk} = [J^{tk}_1, J^{tk}_2, \dots, J^{tk}_N],\] where $J = [x, y, p]$ is a 2D joint coordinate + precision given by the pose estimator. 

Since skeleton coordinates depend on actor camera distance, bounding box dimension and height we normalize each skeleton sequence by subtracting the mid-hip keypoint from each skeleton joint in order to have this last joint as the center of the coordinates system, then we divide each limb by the torso length.

Similarly to~\cite{hcn}, we introduce a representation of skeleton motion. The skeleton motion of a person $k$ at time $t$ is defined as the temporal difference of each joint between two consecutive frames: 
\begin{equation}
\begin{split}
&\mathbf{M}^{tk} = \mathbf{S}^{(t+1)k} - \mathbf{S}^{tk} = \\
&[ J_1^{(t+1)k} - J_1^{tk},J_2^{(t+1)k} - J_2^{tk}, \dots, J^{(t+1)k}_N - J^{tk}_N ]
\end{split}
\end{equation}

\begin{figure*}
	\centering
\includegraphics[width=.9\textwidth]{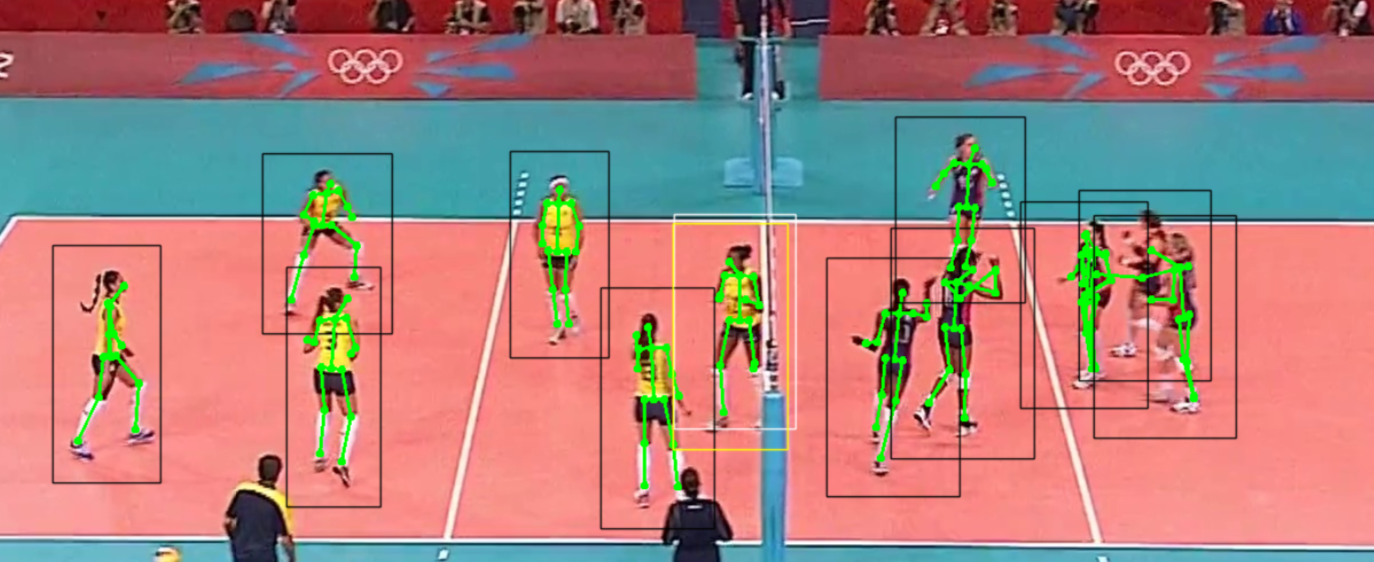}
\caption{\label{fig:pivot} Pivot selection among players. We pick the players closest to the average joint centroid of all players. Pivot is shown in yellow.}
	\end{figure*}

To model the configuration of each actor $k$ we compute a person-to-person interaction $\mathbf{D}^{tk}$ (see Fig.~\ref{fig:pivot}), 
defined as the difference between each pair of joints of two persons at time $t$. In order to obtain a group representation invariant to camera motion and do not rely on global fram coordinates,  we select a pivot actor.  We used the pivot as a  reference to compute the difference between joint pairs. 

We then represent each skeleton sequence via pivot-actor Joint differences, computed between an actor $k$ and the actor pivot $p$ at time $t$ is formulated as: 
\begin{equation}
\mathbf{D}^{tk} = \mathbf{S}^{tk} - \mathbf{S}^{tp}\\
= [ J_1^{tk} - J_1^{tp},J_2^{tk} - J_2^{tp}, \dots, J^{tk}_N - J^{tp}_N ]
\end{equation}
Motion $\mathbf{M}^k$ and pivot-actor joint differences $\mathbf{D}^k$ from all actors are stacked obtaining two tensor $\mathbf{GM}$ and $\mathbf{GD}$ having the same shape of the group skeleton sequence $\mathbf{GS}$.

Group sequence of skeletons, motion and pivot-actor joint differences are fed into the network directly as three input streams into three separate network branches sharing the same architecture. However their parameters are not shared and learned separately. Following~\cite{hcn} feature maps from inputs are learned hierarchically with specific convolution layers, then they are flattened and fused by concatenation obtaining a $F \times K$ matrix to represent feature vectors of actors. 
Feature vectors are used both for action classifications and max-pooled for group activity classification. The whole model can be trained in an end-to-end manner with back-propagation and the extremely small model size allows us to easily train the network from scratch without the need of pretraining. Combining the two standard cross-entropy losses, the final loss function is formed as 
\begin{equation}
	L_{tot} = L_G(y^G,\hat{y}^G) + \lambda L_I(y^I, \hat{y}^I)
\end{equation}
where $L_I$ and $L_G$ are the cross-entropy loss, $y^G$ and $y^I$ denote the ground-truth labels of group activity and individual action, $\hat{y}^G$ and $\hat{y}^I$ are the predictions to group activity and individual action. The first term corresponds to group activity classification loss, and the second is the loss of the individual action classification. The weight $\lambda$ is used to balance these two tasks.

\subsection{Self-Supervised Group Activity Recognition}
The basic idea to provide pseudo-labels for individual actions is to group the representations of persons' crop  into an over-segmented set of clusters, taking the cluster ids as annotations of individual actions. The rationale is that we can exploit the manifold induced by any learned representation to label similar individual actions unsupervisedly.

We use P3D, a pretrained 3D-CNN~\cite{Qiu_2017_ICCV} initialized  on Kinetics sport dataset~\cite{kay2017kinetics} to represent individual actor's skeletal data. We use the last fully connected layer output as features from video clips of each actor. We cluster features from the training set using k-means and use the cluster assignments as "pseudo-labels" $k^I$ to compute individual action loss term $L_I(k^I,\hat{y}^I)$. This allows to train our model in a self-supervised way. Before the clustering, features are PCA-reduced from 2048 to 256 dimensions, whitened and l2-normalized.  

\section{Experiments}
In this section, we evaluate the performance of the proposed method in both supervised and self-supervised variants with several baselines. We also compare our results with the state-of-the-art.

\subsection{Dataset}
We evaluate our method on the Volleyball dataset\cite{msibrahiCVPR16deepactivity}, since it is the only public available dataset for group activity recognition that is relatively large-scale and contains labels for people locations, as well as their collective and individual actions. This dataset contains 4830 clips of 55 volleyball games. Each clip central frame is annotated at each player level with the bounding box and one of the 9 individual actions, and the whole scene is labeled with one of the 8 collective activity. Since other frames are not annotated, to get the bounding boxes of people, we used DLIB tracker\cite{article}. We do horizontal flips as data augmentation.

\subsection{Implementation Details}
We adopt stochastic gradient descent with ADAM to learn the network parameters with fixed hyper-parameters to $\beta_1 = 0.9, \beta_2=0.999, \epsilon=10^{-8}$. We train the network for 100 epochs using a mini-batch size of 64 and a starting learning rate of 0.001 decreasing it by a factor of 10 every 30 epochs. Individual action loss weight $\lambda=0.7$ is used. For training all our models (that include the baseline models) we follow the same training protocol using a Tesla K80 GPU and PyTorch Framework.

\begin{figure}[t]
    \includegraphics[width=0.9\linewidth]{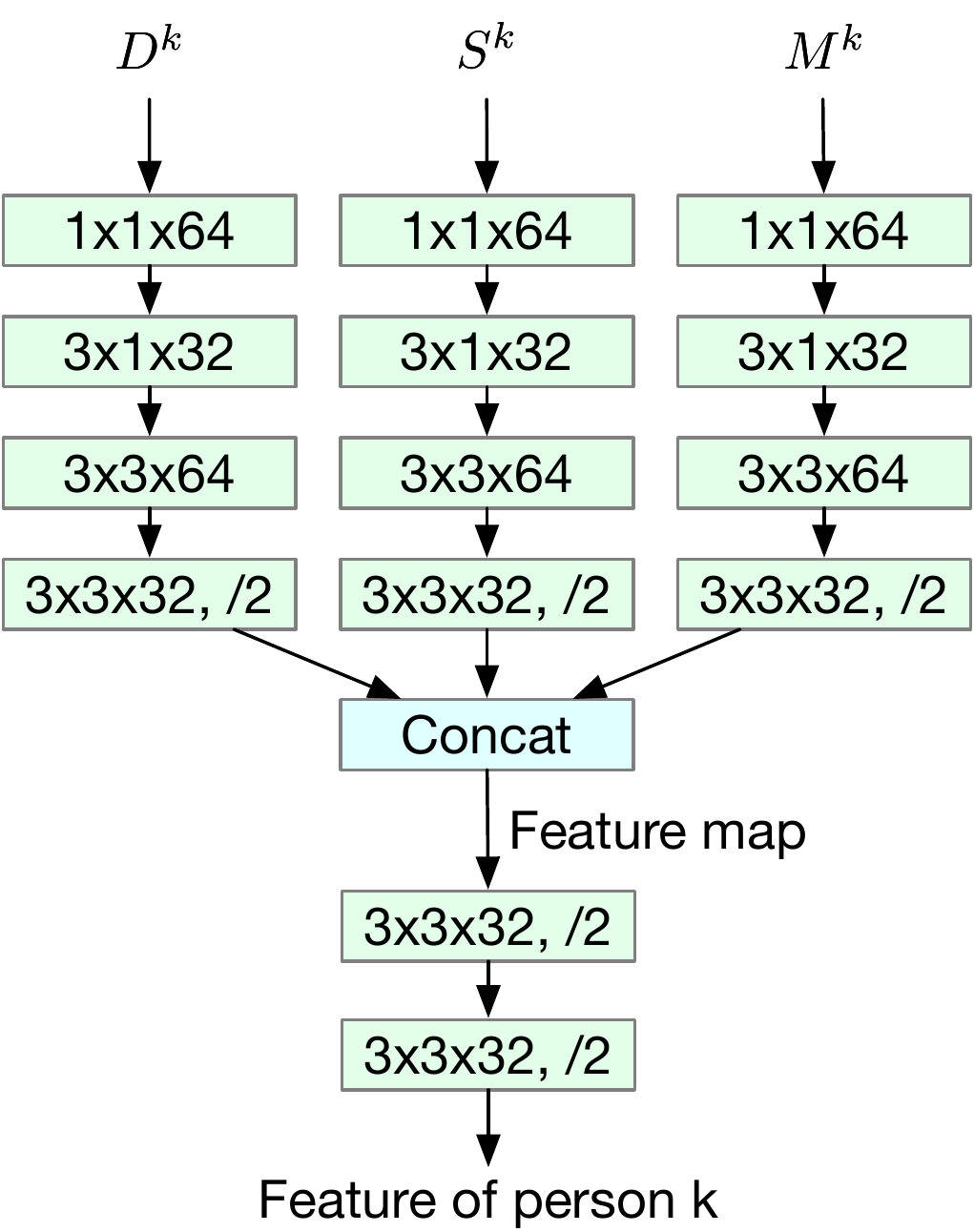}
	\caption{Architecture of the CNN. Each green block corresponds to a convolutional layer of size WxHxFilters. The layers with $/2$ have a stride of 2. The entire figure corresponds to the CNN blocks of Fig. \ref{fig:model}}
	\label{fig:architecture}
\end{figure}

\subsection{Baseline and Variants for Ablation Studies}
Here we evaluate our model by comparing obtained results with several baselines. First, we describe the baseline model then, results on the Volleyball dataset are presented. Our model is end-to-end trainable but could also be implemented in a 2-stage style, splitting the model into an action model and a group model. Training then would consists in learning actions from each skeleton sequence and then use this model to extract actor features, which are pooled over all people and fed to the group model to recognize group activity. 
We test the following approaches:

\newcounter{elenco}

\setcounter{elenco}{0}
\begin{list}{\stepcounter{elenco}\arabic{elenco}}{\setlength{\itemsep}{0.3cm}}
\item \textbf{Two Stage Model without Pivot-Actor Joint Differences}: This baseline is the two stage model with two input streams: skeleton joints and motion. 
\item \textbf{Two Stage Model}: This baseline is an extension of the previous baseline. Here individual action model receive as input a third stream: the pivot-actor joint differences. 
\item \textbf{End-to-end Model without Pivot-Actor Joint Differences}: this baseline is the end-to-end version of first baseline. 
\item \textbf{End-to-end Model}: the end-to-end final version with its three input stream. 
\item \textbf{End-to-end Model with Data Augmentation}: the end-to-end final version with its three input stream and data augmentation.
\end{list}

\begin{table}
    \centering
    \begin{tabular}{l  c } 
    \hline
    Method & Accuracy  \\ [0.5ex] 
    \hline
    Two stage model w/o actor-pivot joint pair difference & 78.1 \\ 
    
    Two stage model & 80.9 \\
    
    End-to-end model w/o actor-pivot joint pair difference & 85.0 \\
    
     End-to-end model & 89.2  \\
    
     End-to-end model with data augmentation & 91.0 \\ 
    \hline
    \end{tabular}
    
\caption{Comparison of our method with baseline methods on the Volleyball dataset.}
\label{tab:baselines}
\end{table}

In Table \ref{tab:baselines}, the classification results of our method is compared with baselines. A performance increase is obtained thanks to Pivot-Actor Joint Differences as it is including a person-to-person context information. End-to-end training helps significantly the model. Also data augmentation is useful for training the model as it provide consistent data for network's learning. Therefore we choose to use both Pivot-Actor Joint Differences and data augmentation with end-to-end training procedure for our model.

\subsection{Self-Supervised Ablation Studies}
As we discussed in previous subsection, the use of Pivot-Actor Joint Differences with end-to-end training is able to achieve the highest performance and we choose this combination as our final model, considering the two variants with and without data augmentation. In order to understand how much performance is affected by the use of our pseudo-labels we conduct the following ablation studies.

\setcounter{elenco}{0}
\begin{list}{\stepcounter{elenco}\arabic{elenco}}{\setlength{\itemsep}{0.3cm}}
\item \textbf{ Group activity labels only}: In this baseline the model has been modified in order to work only with  group activity labels. The layers that receive actor feature representation to classify the individual action are removed and the individual loss factor $\lambda$ is set to zero. This baseline is designed to illustrate the importance of individual activity labels in our model.
\item \textbf{Pseudo action labels from 2D-CNN}: In this baseline we adopted pseudo-labels instead of ground truth action labels. Visual features are extracted from the central frame of each video clip using a pretrained VGG16 2D-CNN. Number of used cluster k is set to 20. This baseline aims to illustrate the importance of pseudo-action-labels.
\item \textbf{ Pseudo action labels from 3D-CNN}: Final Self-Supervised model, where we adopt pseudo-labels instead of ground truth action labels. Visual features are extracted from the whole fixed temporal window of each video clip using a pretrained 3D-CNN. Number of used cluster k is set to 20.
\item \textbf{End-to-end fully supervised}: this method is the end-to-end final version, previously seen in Table \ref{tab:baselines}, trained with the full ground truth. 
\end{list}

The difference in results of the first and second baselines illustrate the importance of using instance label annotation, even if from pseudo-labels. Comparing with the second baseline, our self-supervised method considering the time with a 3D-CNN model obtains better performance. Moreover, our self-supervised method results are very close to the supervised variant, especially when also using data augmentation.

\begin{figure}
	\includegraphics[width=\columnwidth]{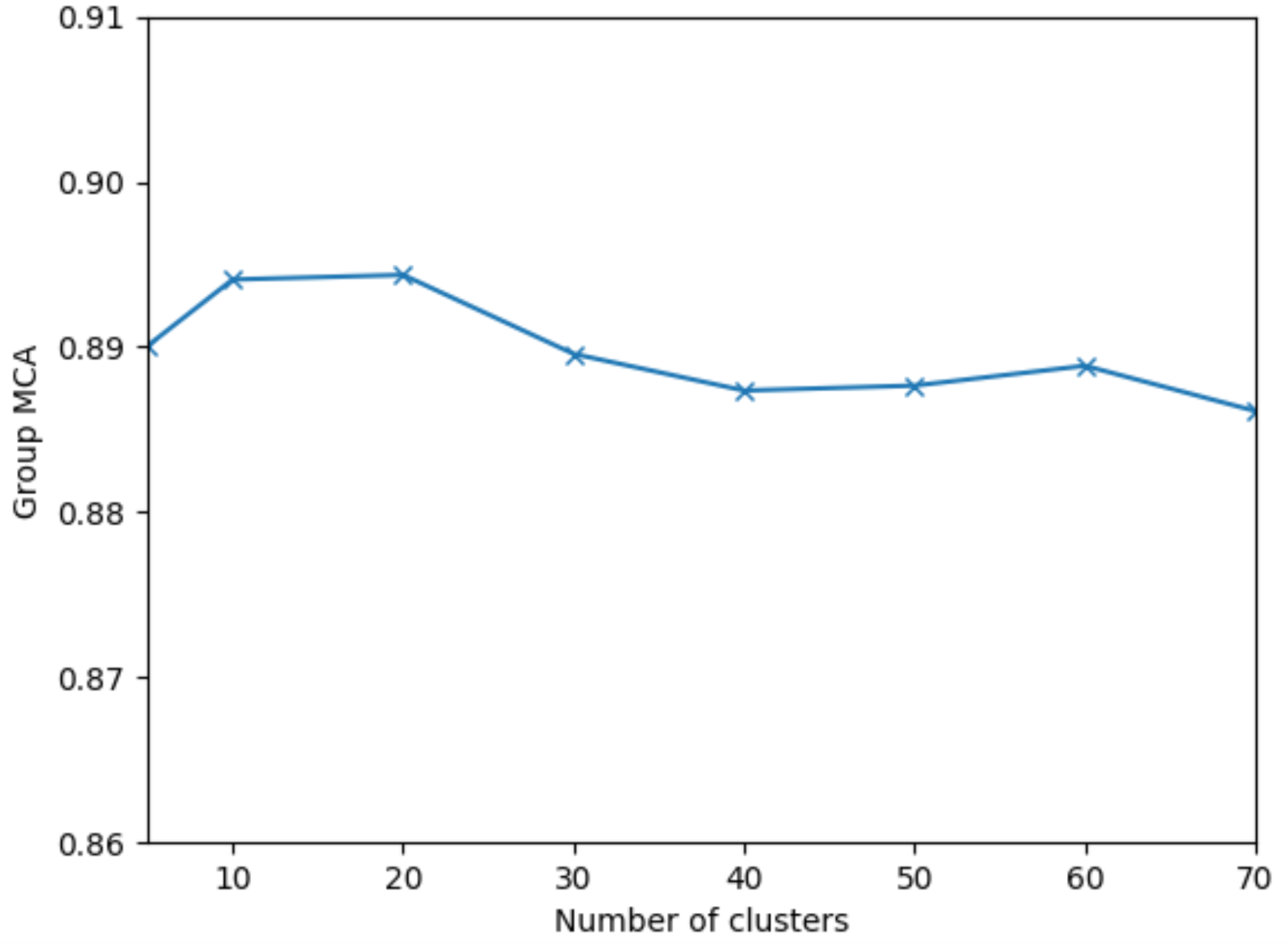}
	\caption{\label{fig:kmeans} Mean Classification Accuracy of group activity varying the number of clusters for self-supervised learning. }
\end{figure}
Fig.~\ref{fig:kmeans} shows group accuracy results obtained by varying the number of clusters by a step of 10, note that best results are obtained with k=20. Given that we train our model on the Volleyball dataset, one would expect k=9 (actual number of action classes) to yield the best results, but apparently some amount of over-segmentation is beneficial.


\subsection{Comparison with state-of-the-art methods}
There are only a few works~\cite{azar2018multi},~\cite{cvpr2020transformer}  using pose in group activity recognition reporting results on the Volleyball dataset, thus we compare our method that exploits only poses with other state of-the-art methods that make use of different input feature like RGB and/or optical flow.  As shown in Table \ref{tab:sota_comparison}, our method is very competitive with the state-of-the-art methods and even outperforms most of the methods that exploit RGB and optical flow input (including PC-TDM~\cite{yan2018participation} and SPA+KD+OF~\cite{tang2019learning}). Although ARG~\cite{wu2019learning}, CRM~\cite{mokhtarzadeh2019convolutional}, PRL~\cite{hu2019progressive} and AT~\cite{cvpr2020transformer} perform somewhat better than our method, note that it is unfair to compare with, because they use RGB, optical flow input  and  also  a  much  larger  model  than  ours.   In  addition,  even  our  self-supervised  action  method  outperformed  various  approaches  without using ground truth  individual  action  labels, showing that it is possible to obtain good results using  action labels generated with visual feature clustering. 

\begin{table}
    \centering
    \begin{tabular}{ l  c c } 
    \hline
    Method & No Data Aug. & With Data Aug. \\ [0.5ex] 
    \hline
    Group activity labels only& 84.9 & 87.1 \\ 
    Pseudo action labels from 2D-Vgg16 & 87.2 & 89.3 \\
    Pseudo action labels from 3D-Resnet & 87.5 & 89.5 \\
    Supervised & 89.2 & 91.0 \\
    \hline
    
    \end{tabular}
    
\caption{Comparison of our method with and without action labels. SSAL stands for Self Supervised Action Learning corresponding to centroid indexes assigned by clustering of visual features.}
\label{tab:selfsupervised_baselines}
\end{table}
\begin{table}
    \centering
    \begin{tabular}{ l l c r } 
    \hline
    Method & Input & Action Labels & Accuracy \\ [0.5ex] 
    \hline
    HDTM \cite{msibrahiCVPR16deepactivity} & RGB & Yes & 81.9 \\ 
    SSU \cite{bagautdinov2017social} & RGB & Yes & 89.9 \\
    PC-TDM \cite{yan2018participation} & RGB+OF & Yes & 87.7 \\
    MS-CNN \cite{azar2018multi} & RGB+POSE & Yes & 90.5 \\
    stagNet \cite{qi2018stagnet} & RGB & Yes & 89.3 \\
    RCRG \cite{ibrahim2018hierarchical} & RGB & Yes & 89.5 \\
    SPA+KD \cite{tang2019learning} & RGB & Yes & 89.3 \\
    SPA+KD+OF \cite{tang2019learning} & RGB+OF & Yes & 90.7 \\
    ARG \cite{wu2019learning} & RGB & Yes & 92.6 \\
    CRM \cite{mokhtarzadeh2019convolutional} & RGB+OF & Yes & 93.0 \\
    PRL \cite{hu2019progressive}& RGB & Yes & 91.4 \\
    AT \cite{cvpr2020transformer}& POSE+OF & Yes & 94.4 \\
    \hline
    Ours-SSAL & POSE & No & 89.4 \\
    Ours & POSE & Yes & 91.0 \\
    \hline
    \end{tabular}
    
\caption{Comparison of recognition accuracy (\%) on Volleyball dataset. "OF" denotes optical flow input, while column "Action Labels" says that a method use ground truth annotations at individual actor level.}
\label{tab:sota_comparison}
\end{table}

\section{Conclusion}
In this work we have shown a solution to train group activity recognition systems end-to-end without the use of individual action labels. Our method using only skeletal data is able to reach state-of-the art performance without using RGB or Optical flow and requiring only labels at the sequence level. Compared to previous work, the proposed approach needs less supervision to be trained and using only skeleton data, can also be used in contexts where privacy require to avoid saving RGB images.







\bibliographystyle{IEEEtran}
\bibliography{IEEEabrv,references}
%

\end{document}